\documentclass[runningheads]{llncs}
\usepackage{graphicx}
\usepackage{dsfont}
\usepackage{amsmath}
\usepackage{float}
\usepackage{mathtools}
\usepackage{amsmath}
\usepackage{caption}

\usepackage{subcaption}
\usepackage{textalpha}
\usepackage{dsfont}
\usepackage{amsmath}
\usepackage{float}
\usepackage{multirow}
\usepackage{hhline}
\usepackage{longtable}
\usepackage{caption}
\usepackage{xcolor,pifont}
\usepackage{multicol}
\usepackage[left=38.5mm,right=38.5mm,top=24.5mm,bottom=33.95mm]{geometry}
\usepackage{tabularx}
\usepackage{amssymb}
\usepackage{pifont}
\usepackage{adjustbox}
\usepackage{lipsum}
\usepackage{tikz}
\usepackage{enumitem}
\usepackage{booktabs}
\usepackage{xcolor,pifont}
\usepackage{amssymb}
\usepackage{pifont}
\newcommand*\colourcheck[1]{%
  \expandafter\newcommand\csname #1check\endcsname{\textcolor{#1}{\ding{52}}}%
}

\usepackage[colorlinks=true, citecolor=citecolor, linkcolor=linkcolor]{hyperref}
\definecolor{citecolor}{HTML}{0071BC}
\definecolor{linkcolor}{HTML}{ED1C24}
\usepackage{verbatim}
\usepackage[capitalize]{cleveref}

\begin{document}

\title{Towards End-to-End Semi-Supervised Table Detection with Semantic Aligned Matching Transformer}
\titlerunning{Semi-Supervised Table Detection with Semantic Aligned Matching Transformer}
\author{Tahira Shehzadi*\inst{1,2,3}\orcidID{0000-0002-7052-979X} \and
Shalini Sarode\inst{1,3}\orcidID{0009-0007-9968-4068} \and \\
Didier Stricker\inst{1,2,3} \and
Muhammad Zeshan Afzal\inst{1,2,3}\orcidID{0000-0002-0536-6867}}

\authorrunning{T. Shehzadi et al.}

\institute{Department of Computer Science, Technical University of Kaiserslautern, 67663, Germany \and
Mindgarage, Technical University of Kaiserslautern, 67663, Germany \and
German Research Institute for Artificial Intelligence (DFKI), 67663 Kaiserslautern, Germany\\
\email{\{tahira.shehzadi@dfki.de\}}
}
\maketitle              

\begin{abstract}
Table detection within document images is a crucial task in document processing, involving the identification and localization of tables. Recent strides in deep learning have substantially improved the accuracy of this task, but it still heavily relies on large labeled datasets for effective training. Several semi-supervised approaches have emerged to overcome this challenge, often employing CNN-based detectors with anchor proposals and post-processing techniques like non-maximal suppression (NMS). However, recent advancements in the field have shifted the focus towards transformer-based techniques, eliminating the need for NMS and emphasizing object queries and attention mechanisms. Previous research has focused on two key areas to improve transformer-based detectors: refining the quality of object queries and optimizing attention mechanisms. However, increasing object queries can introduce redundancy, while adjustments to the attention mechanism can increase complexity. To address these challenges, we introduce a semi-supervised approach employing SAM-DETR, a novel approach for precise alignment between object queries and target features. Our approach demonstrates remarkable reductions in false positives and substantial enhancements in table detection performance, particularly in complex documents characterized by diverse table structures. This work provides more efficient and accurate table detection in semi-supervised settings.
\end{abstract}

\section{Introduction}
Document analysis has been the fundamental task in various workflow pipelines\cite{Breuel2017,articlew}, with document summarization as its core task. The essential task in document analysis is identifying graphical objects like tables, figures, and text paragraphs. Previously, this task was carried out manually by analyzing the documents, understanding their contents, and summarizing them. However, the number of documents that need to be analyzed has drastically increased, and manual inspection is impossible. The growing number of documents led businesses to use more efficient and reliable automated methods. Optical character recognition(OCR)~\cite{ocr8,ocr4} and rule-based table detection approaches\cite{RecogTable5,TTsurvey8,extractTab9} are classical approaches for visual summarization. These methods perform well for documents with highly structured layouts because they are rule-based\cite{RecogTable5,TTsurvey8,extractTab9}. However, they struggle to adapt to varying and newer table designs, such as borderless tables. These limitations has shifted the research focus to developing techniques using deep learning \cite{fast15,faster23,YOLO9000,mask-rcnn84}. These methods show significant improvements over traditional approaches \cite{mlr7}, precisely detecting tables in documents irrespective of their structure. This advancement provides a notable improvement in document analysis and visual summarization.

Deep learning methods \cite{DeepDeSRT3,TDcl34,TSRkhurm4,cas10,rethink78,naik86} eliminate handcrafted rules and excel at generalizing problems. However, their reliance on large amounts of labeled data for training counteracts the aim of reducing manual work. Generating these labels is time-consuming and prone to errors~\cite{inbook}. Although these supervised deep learning approaches achieve state-of-the-art results on public benchmarks, their usage in industries is limited without similarly large annotated datasets in specific domains. 
Semi-supervised learning methods~\cite{van2020survey} have emerged as a solution to insufficient labeled data for deep learning applications. Recent advancements~\cite{omnisup8,rethink98,selfsup6} utilize two detectors: one generates pseudo-labels for unlabeled data, and the other refines predictions using these pseudo-labels and a smaller set of labeled data. These detectors update each other throughout training \cite{selfsup87,propsemi6,activsemi3,unsupaug4}. However, it's important to note that the initial pseudo-label generator is often not robust, potentially leading to inaccurate labels and affecting overall performance.

Additionally, there are two major drawbacks in the earlier CNN-based semi-supervised methods\cite{softTeacher56,omnisup8,rethink98}: First, they rely on anchor points for region proposals that require manual tuning. Second, they use post-processing techniques like Non-Maximal Suppression(NMS) to limit the number of overlapping predictions.
The emergence of transformer-based methods \cite{dino23,Deformable54,adamixer7,co_detrs} make the network end-to-end without NMS and anchor-free. This is possible due to their dependence on the attention mechanism and object queries. Consequently, there has been research mainly to improve the quality of object queries and improve the attention mechanism\cite{shehzadi2023object}. For example, Deformable DETR~\cite{Deformable54}, AdaMixer~\cite{adamixer7} and REGO~\cite{rego2} focus on advancing the attention mechanism. Meanwhile, models like DN DETR~\cite{dn42}, DAB DETR~\cite{dab89}, and DINO DETR~\cite{dino23} are dedicated to improving the quality of object queries, and H-DETR~\cite{hdetrs}, Co-DETR~\cite{co_detrs}, and FANet~\cite{FNet45} aim to increase the quantity of object queries. However, this increase leads to redundant predictions, adversely affecting performance. To counter this, a dual-stage object query approach has been proposed, combining one-to-one and one-to-many matching strategies. Despite its effectiveness, this method still impacts performance~\cite{hdetrs}. Addressing these challenges, we employ SAM-DETR~\cite{sm-detr34}, a novel model designed to optimize the matching process between object queries and corresponding target features in a semi-supervised setting. This approach effectively reduces false positives and improves table detection performance in complex documents.

In this paper, we introduce a novel semi-supervised approach for table detection, employing SAM-DETR~\cite{sm-detr34} detector. Our main objective is to solve the non-robustness of the pseudo-label generation process. The training procedure consists of two modules: the teacher and the student. The teacher module consists of a pseudo-labeling framework, and the student uses these pseudo-labels along with a smaller set of labeled data to produce the final predictions. The pseudo-labeling process is optimized by iteratively refining the labels and the detector. The teacher module is updated by an Exponential Moving Average (EMA) from the student to improve the pseudo-label generation and detection modules. Our approach differs from conventional pseudo-labeling methods by incorporating a SAM-DETR detector without object proposal generation and post-processing steps like NMS. We enhance the ability to accurately match object queries with corresponding target features in complex documents, particularly excelling in the detection and handling of tables in semi-supervised settings. The intrinsic flexibility of this method enables consistent and reliable performance in various scenarios, including diverse table sizes and scales, within a semi-supervised learning context. Furthermore, this framework creates a reinforcing loop where the Teacher model consistently guides and improves the Student model. Our evaluation results demonstrate that our semi-supervised table detection approach achieves superior results compared to both CNN-based and other transformer-based semi-supervised methods without needing object proposals and post-processing steps such as NMS.

\noindent We summarize the primary contributions of this paper as follows:
\begin{itemize}
    \item[$\bullet$] We introduce a novel semi-supervised approach for table detection. This approach eliminates the need for object proposals and post-processing techniques like Non-maximal Suppression (NMS).
    \item[$\bullet$] To the best of our knowledge, this is the first network that optimizes the matching process between object queries and corresponding target features in a semi-supervised setting.
    \item[$\bullet$] We conduct comprehensive evaluations on four diverse datasets: PubLayNet, ICDAR-19, TableBank, and Pubtables. Our approach achieves results comparable to CNN-based and transformer-based semi-supervised methods without requiring object proposal processes and Non-maximal Suppression (NMS) in post-processing.
\end{itemize}

\section{Related Work}
\label{sec:Literature-Review}
Analyzing document images involves the integral table detection task. This segment summarizes techniques for detecting tables, especially those involving complex structures. Initial methods relied on rules or metadata \cite{tsk5,tupextract3,strctRTB3,DPmatch4}. Meanwhile, more recent advances employ statistical and deep learning techniques \cite{DeepDeSRT3,DeCNT82,CasTab45,Hyb65}, improving system adaptability and generalizability.

\subsection{Table Detection Approaches}
\noindent\textbf{Rule-based Approaches}
Itonori et al. \cite{tsk5} laid the groundwork for table detection. The central focus was identifying tables as distinct text blocks using predefined rules. Building upon this, methods like \cite{strctRTB3} improved the approach by integrating various techniques, including table detection based on layout \cite{TINTIN67} or extracting tables from HTML-formatted documents \cite{PIV67}. Although effective for specific document types, these rule-based methods\cite{RecogTable5,TTsurvey8,extractTab9,DEA38,TSsurvey32} lacked the flexibility to be universally applicable.

\noindent\textbf{Learning-based Approaches}
Cesarini et al. \cite{trainTD5} deviates from rule-based approaches by pioneering a supervised learning system for identifying table objects in document images. Their approach transforms a document image into an MXY tree model by classifying the blocks surrounded by vertical and horizontal lines as table objects. They further employed  Hidden Markov Models \cite{RichMM3,FTabA4} and an SVM classifier, along with conventional heuristics \cite{lineln6} for table detection. These techniques still needed additional data like ruling lines. In contrast, Deep Learning-based methods, further categorized as object detection, semantic segmentation, and bottom-up approaches, have demonstrated superior accuracy and efficiency over traditional techniques.

\noindent\textbf{Approaches Based on Semantic Segmentation.} 
Approaching table detection as a segmentation problem, methods like \cite{Xi17,He761,Ik36,Paliw9} generate pixel-level segmentation masks and then aggregate the masks to achieve final table detection. These methods utilize existing semantic segmentation networks and outperform traditional methods on various benchmark datasets \cite{icdar19,PubLayNet3,iiit13k,icdar13,icdar17,tablebank8,pubtables5}. Yang et al.'s \cite{Xi17} approach introduced a fully convolutional network (FCN) \cite{FCNseg4}. They used additional linguistic and visual features to enhance the segmentation results of page objects. He et al. \cite{He761} developed a multi-scale FCN that generates segmentation masks and their contours for table/text areas. They isolate the final table blocks after further refining the masks.


\noindent\textbf{Bottom-Up Methods.}
These methods treat table detection as a graph-labeling task with graph nodes as page elements and edges as connections between them. Li et al. \cite{Li64} used a conventional layout analysis to identify line areas. They then utilized two CNN-CRF networks to categorize these lines into four classes: text, figure, formula, and table. Later, they predicted a cluster for each pair of line areas. Holecek et al. \cite{Martin66} and Riba et al. \cite{busmessg4} constructed a graph to establish the document layout and viewed text areas as nodes. They then used graph-neural networks for classifying nodes and edges. These methods require certain assumptions, like the necessity of text line boxes as additional input.


\noindent\textbf{Object Detection-Focused Techniques}
Table detection in document images~\cite{continuaLR45,Real_DICls4} is considered an object detection challenge, treating tables as natural objects. Hao et al.~\cite{Hao789} and Yi et al.~\cite{Yi77} utilized R-CNN for table detection, but their performance still depended on heuristic rules, similar to earlier methods. Subsequently, more advanced single-stage object detectors like RetinaNet~\cite{retinaNet68} and YOLO \cite{yolos6}, as well as two-stage detectors like Fast R-CNN~\cite{fast15}, Faster R-CNN \cite{faster23}, Mask R-CNN~\cite{mask86}, and Cascade Mask R-CNN~\cite{cascadercnn8}, were employed for detecting various document elements, including figures and formulas~\cite{Pog84,Azka62,yolotab5,gte9,Ayan29,Agarwal52,DeepDeSRT3,bridging_per8}. Additional enhancement techniques, such as image transformations involving coloration and dilation, were applied by \cite{Azka62,Ayan29,arif48}. Siddiqui et al.~\cite{Sidd32} integrate deformable convolution and RoI-Pooling~\cite{deformconv4} into Faster R-CNN for improved handling of geometrical changes. Agarwal et al.~\cite{Agarwal52} combined a composite network~\cite{CBNet5} with deformable convolution to enhance the efficiency of the two-stage Cascade R-CNN. These CNN-based object detectors include heuristic stages like proposal generation and post-processing steps like non-maximal suppression (NMS). Our semi-supervised model treats detection as a set prediction task, eliminating the need for anchor generation and post-processing stages like NMS, resulting in a more streamlined and efficient detection process.

\subsection{Semi-Supervised Learning in Object Detection}
Semi-supervised object detection can be classified into consistency-based methods~\cite{consemi6,propsemi8} and pseudo-label generation methods~\cite{omnisup8,rethink98,selfsup6,semimask1,Ksemi76,sparse_semi_detr1,simplesemi76,minsemi8}. Our work focuses on the latter. Earlier works~\cite{omnisup8,rethink98} employ diverse data augmentation techniques to generate pseudo-labels for unlabeled data. Meanwhile,~\cite{selfsup6} introduces SelectiveNet for pseudo-label generation by superimposing a bounding box from an unlabeled image onto a labeled image to ensure localization consistency within the labeled dataset. However, this approach involves a complex detection process due to image alteration. STAC~\cite{simplesemi76} proposes to use strong augmentation for pseudo-label creation and weak augmentation for model training. Our method introduces a seamless end-to-end semi-supervised approach for table detection. Similar to other pseudo-label techniques~\cite{omnisup8,rethink98,selfsup6,simplesemi76,minsemi8}, it incorporates a multi-level training strategy without the need for anchor generation and post-processing steps like Non-Maximal Suppression (NMS).

\section{Methodology}
\label{sec:method}
First, the paper reviews SAM-DETR, a recent approach for detecting objects using transformers, in Section~\ref{sec:sam-detr}. Then, Section~\ref{sec:semi-sup} describes our semi-supervised approach for learning with limited supervision and the generation of pseudo-labels for training.

\subsection{Revisiting SAM-DETR}
\label{sec:sam-detr}
\begin{figure*}
\centering
\includegraphics[width=0.8\textwidth]{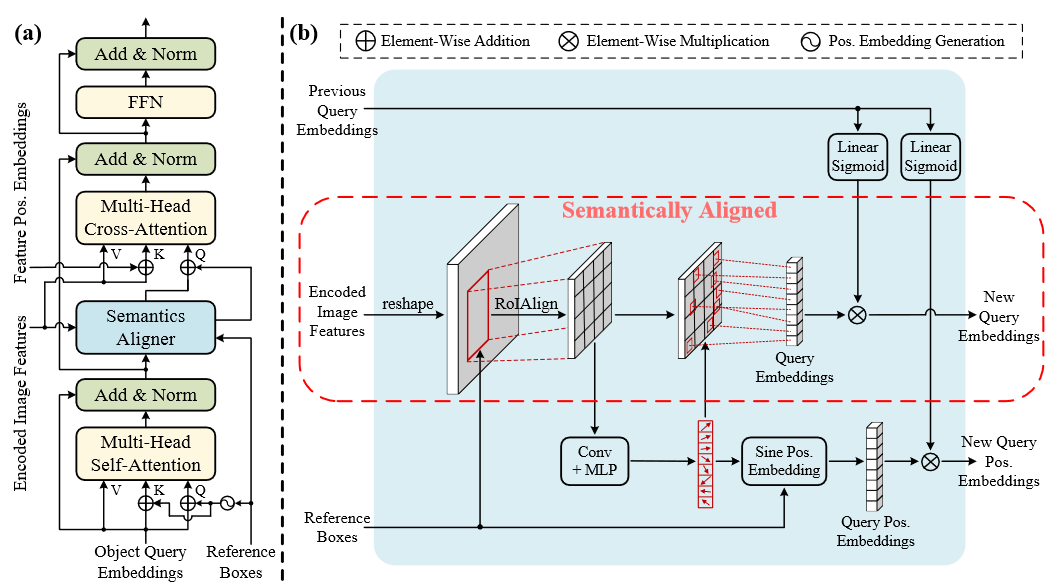}
\caption{Overview of SAM-DETR~\cite{sm-detr34}. (a) the architecture of a single decoder layer in SAM-DETR, showing the role of learnable reference boxes in generating position embeddings for each object query. (b) the pipeline of the Semantics Aligner. The process includes the use of reference boxes for feature extraction via RoIAlign, the prediction of salient points in the targeted region, and the generation of new, semantically aligned query embeddings, which are further refined by incorporating attributes from previous queries. Image from~\cite{sm-detr34}. }\label{fig:encoder-decoder}
\end{figure*}

DEtection TRansformer (DETR)~\cite{detr34} introduces an encoder-decoder network for object detection. The encoder network extracts features from the image to focus on key details. The decoder then processes these features with object queries, using self-attention and cross-attention mechanisms to identify and locate objects. However, DETR's initial non-selective approach in processing images and object queries can lead to slower detection, especially in semi-supervised learning with limited data. By refining the attention mechanism and enhancing the quality and quantity of object queries, researchers aim to boost DETR's efficiency, accuracy, and training speed~\cite{shehzadi2023object}. SAM-DETR, as shown in Fig.~\ref{fig:encoder-decoder} stands out for its innovative addition of a semantics aligner module and learnable reference boxes within the Transformer decoder part of DETR. Overall, SAM-DETR's enhancements to the original DETR model focus on making the object detection process more efficient in terms of accuracy and speed.

\noindent\textbf{Semantics Aligner.}
Semantic-Aligned Matching focuses on improving the interaction between object queries and encoded image features. Generally, the cross-attention module uses a dot-product method, which is effective in identifying similarities between two vectors. This method typically guides object queries to focus on regions of the image that are more similar. However, the original DETR model does not ensure that object queries and encoded image features are in the same embedding space, leading to less effective matching and requiring extensive training time. To address this, the Semantic-Aligned Matching approach introduces a mechanism to align object queries with encoded image features semantically. This alignment ensures that both are in the same embedding space, making the dot-product a more meaningful measure of similarity. As a result, object queries are more likely to focus on semantically similar regions, enhancing the efficiency and effectiveness of the object detection process.

\noindent\textbf{Multi-Head Attention and Salient Points.}
In DETR, multi-head attention is crucial for focusing on different image parts, enhancing scene understanding. SAM-DETR builds on this by identifying key points on objects, using ConvNet and MLP to predict these points for better alignment and detection. Features from these points are integrated with multi-head attention, allowing each head to concentrate on specific, significant object features, improving accuracy and localization.

\noindent\textbf{Reweighted Queries.}
The Semantics Aligner in DETR aligns object queries with encoded image features but initially misses crucial information from previous embeddings. To address this, it uses a linear projection and sigmoid function to create reweighting coefficients, applied to both new and positional query embeddings. This ensures important features are emphasized and previous data is utilized, significantly enhancing detection.
\begin{figure}
\centering
\includegraphics[width=0.8\textwidth]{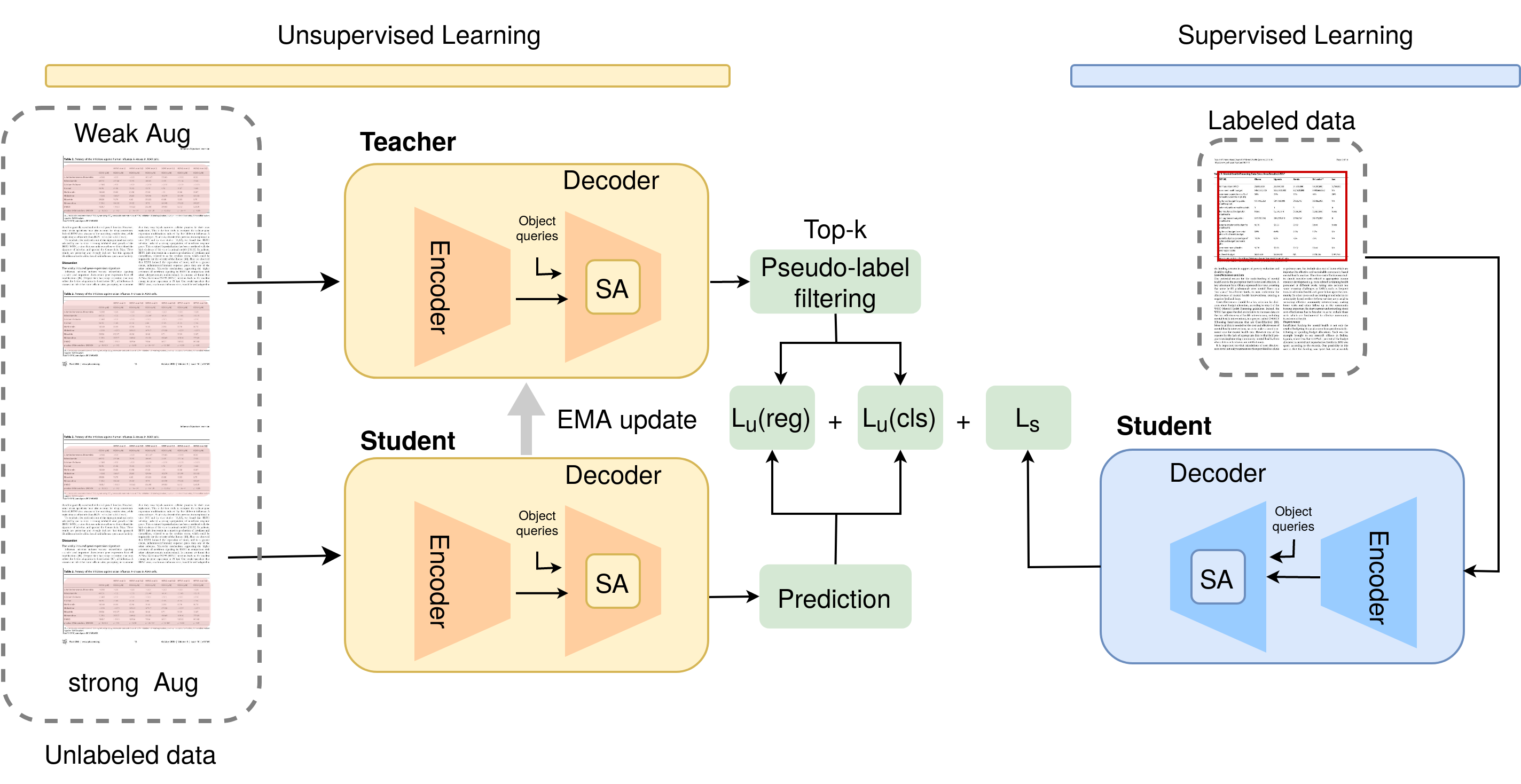}
\caption{Illustration of our Semi-Supervised Table Detection Framework. This dual-component system involves a Student module that learns from a mix of labeled data and strongly augmented unlabeled images, and a Teacher module that refines its understanding using weakly augmented unlabeled images. The Student module updates the Teacher module using Exponential Moving-Average (EMA) during training. Within this setup, the Semantics Aligner (SA) is key in the decoder of the student-teacher framework, fine-tuning the relationship between object queries and the image features that have been encoded, ensuring a more effective and accurate detection of tables in various documents.}\label{fig:semi}
\end{figure}
\subsection{Semi-Supervised SAM-DETR }
\label{sec:semi-sup}
We propose a semi-supervised learning approach that improves object detection through semantic alignment and utilizes limited labeled data for training, as shown in Fig.~\ref{fig:semi}. The model leverages fully labeled and unlabeled data for object detection tasks in the semi-supervised setting. It consists of two key modules: the student and teacher modules. The student module processes both labeled and unlabeled images. Strong augmentation is applied to unlabeled data, while strong and weak augmentations are applied to labeled data. The teacher module operates on unlabeled images with weak augmentations. It plays a crucial role in generating pseudo-labels for unlabeled data. These pseudo-labels are then employed for supervised training by the student module. Weak augmentation is applied to the unlabeled data for the teacher module to produce more accurate pseudo-labels. In contrast, the student module, designed for more challenging learning, utilizes strong augmentation for unlabeled data. At the start of training, the teacher and student models are randomly initialized. As training progresses, the teacher model is continuously updated by the student model using an exponential moving average (EMA) strategy.
For the student module, the student's queries $Q_s$ and features $F_s$ are fed into the decoder. Similarly, in the teacher module, the teacher's queries $Q_t$ and features $F_t$ go through a similar process with the teacher's decoder as follows:
\begin{equation}
\hat{o}_s = \text{Decoder}_s\left(Q_s, F_s \right)
\end{equation}
\begin{equation}
\hat{o}_t = \text{Decoder}_t\left(Q_t, F_t \right)
\end{equation}
In the decoder, the Semantics Aligner processes the encoded image features for students \( F_s \) and teachers \( F_t \), both initially in 1D sequences of dimensions \( HW \times d \). The Aligner converts these features into 2D maps with dimensions \( H \times W \times d \), using the reference boxes of object queries, denoted as \( R_s^{box} \) for the student and \( R_t^{box} \) for the teacher. After this transformation, the aligner employs RoIAlign to extract region-level features, represented as \( F_s^R \) for the student and \( F_t^R \) for the teacher, from the encoded image features. The final step involves generating new object queries, \( Q_{\text{new}} \) and their position embeddings\( Q_{\text{new pos}} \), through resampling based on \( F_s^R \) and \( F_t^R \) as follows.
\begin{equation}
F_s^R = \text{RoIAlign}(F_s, R^{\text{box}}_s), \hspace{1em} F_t^R = \text{RoIAlign}(F_t, R^{\text{box}}_t)
\end{equation}
\begin{equation}
Q_s^{\text{new}}, Q^{\text{new}}_{\text{s,pos}} = \text{Resample}(F_s^R, R_s^{\text{box}}, Q_s), 
\end{equation}
\begin{equation}
Q_t^{\text{new}}, Q^{\text{new}}_{\text{t,pos}} = \text{Resample}(F_t^R, R_t^{\text{box}}, Q_t)
\end{equation}
Next, we extract features via a ConvNet and MLP to identify salient points within these regions. These points are then used to create new object query embeddings $Q_s^{\text{new}}$ and $Q_t^{\text{new}}$, ensuring they stay within reference boxes for accuracy. Finally, position embeddings $Q^{\text{new}}_{\text{s,pos}}$ and $Q^{\text{new}}_{\text{t,pos}}$ derived from these points are concatenated, feeding into a multi-head cross-attention module for further processing.

\begin{equation}
R^{sp}_s = MLP(ConvNet(F^R_s))
\end{equation}
\begin{equation}
Q_s^{\text{new}} = \text{Concat}\left(\left\{F_s^R[\ldots, x, y, \ldots] \text{ for } x, y \in R_s^{\text{sp}}\right\}\right)
\end{equation}
\begin{equation}
Q^{\text{new}}_{\text{s,pos}} = \text{Concat}(\text{Sin}(R_s^{\text{box}}, R_s^{\text{sp}}))
\end{equation}

\begin{equation}
R^{sp}_t = MLP(ConvNet(F^R_t))
\end{equation}
\begin{equation}
Q_t^{\text{new}} = \text{Concat}\left(\left\{F_t^R[\ldots, x, y, \ldots] \text{ for } x, y \in R_t^{\text{sp}}\right\}\right)
\end{equation}
\begin{equation}
Q^{\text{new}}_{\text{t,pos}} = \text{Concat}(\text{Sin}(R_t^{\text{box}}, R_t^{\text{sp}}))
\end{equation}
The semantics aligner generates new object queries aligned with image features and incorporates previous query embeddings by generating reweighting coefficients. These coefficients, created through linear projection and sigmoid functions, are applied to new and old query embeddings to emphasize key features. This approach ensures that the valuable information from previous queries is effectively utilized.
\begin{equation}
Q_s^{\text{new}} = Q_s^{\text{new}} \otimes \sigma({Q_sW}_s^{\text{RWs1}}), \hspace{4em} Q_t^{\text{new}} = Q_t^{\text{new}} \otimes \sigma({Q_tW}_t^{\text{RWt1}}) 
\end{equation}
\begin{equation}
Q^{\text{new}}_{\text{s,pos}} = Q^{\text{new}}_{\text{s,pos}} \otimes \sigma({Q_sW_s}^{\text{RWs2}}), \hspace{3em} Q^{\text{new}}_{\text{t,pos}} = Q^{\text{new}}_{\text{t,pos}} \otimes \sigma({Q_tW_t}^{\text{RWt2}})  
\end{equation}
Here, \( W_{\text{RWt1}} \) and \( W_{\text{RWt2}} \) are used to denote linear projection functions. The symbol \( \sigma(\cdot) \) refers to the sigmoid function, while \( \otimes \) represents the operation of element-wise multiplication. The subscripts t and s refer to the teacher and student module, respectively. Combining the semantic alignment capabilities with the semi-supervised approach allows the model to effectively utilize labeled and unlabeled data, leading to improved object detection performance. This approach is particularly useful when labeled data is limited, as it maximizes the information extracted from available resources.

\section{Pseudo-Label Filtering Framework}
In our semi-supervised learning framework, we employ the Top-K pseudo-label filtering technique to augment the training process of our machine learning models, especially when the labeled data is limited. This approach is instrumental in making the most of the unlabeled data. Here, the key strategy is pseudo-labeling, where our model generates labels for the unlabeled data based on its current level of understanding. However, diverging from the traditional method of relying on the single most confident prediction, our top-k approach considers each data point's top 'k' predictions. For instance, if 'k' is set at 3, the model evaluates and includes the three highest probable labels for each piece of unlabeled data in the training process. The benefits of our top-k strategy are significant. Firstly, it broadens the model's exposure to more challenging 'hard samples' data points that are typically difficult to classify and might be overlooked by standard top-1 pseudo-labeling methods. Including a wider range of examples substantially improves the model's learning. Secondly, our approach is effective in cases involving objects or data points with similar features. By acknowledging and incorporating ambiguity through multiple potential labels, the model is better equipped to handle complex classification scenarios where clear-cut distinctions between categories are not always evident. Implementing the top-k pseudo-label filtering in our semi-supervised learning setting is a pivotal step towards enhancing the model's accuracy and robustness, ensuring a more comprehensive and enhanced learning process.
The teacher model generates pseudo boxes for unlabeled images, and the student model is trained on labeled images with ground-truth annotations and unlabeled images with pseudo boxes treated as ground-truth. Therefore, the overall loss is defined as the weighted sum of supervised and unsupervised losses:

\begin{equation}
L = L_s + \alpha L_u, \quad 
\end{equation}

Where \(L_s\) represents the supervised loss for labeled images, \(L_u\) represents the unsupervised loss for unlabeled images, and \(\alpha\) with value 0.25 controls the contribution of the unsupervised loss. Both losses are normalized by the respective number of images in the training data batch:
\begin{equation}
L_s = \frac{1}{N_l} \sum_{i=1}^{N_l} (L_{cls}(I_{i}^{l}) + L_{reg}(I_{i}^{l})), \quad 
\end{equation}

\begin{equation}
L_u = \frac{1}{N_u} \sum_{i=1}^{N_u} (L_{cls}(I_{i}^{u}) + L_{reg}(I_{i}^{u})), \quad 
\end{equation}
Where \(I_{i}^{l}\) indicates the \(i\)-th labeled image, \(I_{i}^{u}\) indicates the \(i\)-th unlabeled image, \(L_{cls}\) is the classification loss, \(L_{reg}\) is the box regression loss, \(N_l\) is the number of labeled images, and \(N_u\) is the number of unlabeled images. Overall, our semi-supervised learning setting enhances the model's accuracy and robustness, ensuring a more comprehensive learning process.
\section{Experimental Setup}
\label{sec:exp}
\subsection{Datasets}
\label{sec:dataset}
\noindent\textbf{TableBank:} TableBank \cite{tablebank8}, a prominent dataset in the field of document analysis, ranks as the second-largest collection for table recognition tasks. This dataset comprises 417,000 document images, annotated via a process of crawling the arXiv database. It categorizes tables into three types: LaTeX images (253,817), Word images (163,417), and a combined set (417,234). Furthermore, TableBank provides data for table structure recognition. In our study, we utilizeonly the table detection component of the TableBank dataset.

\noindent\textbf{PubLayNet:}
PubLayNet \cite{PubLayNet3}, a sizable dataset in the public domain, encompasses 335,703 images for training, 11,240 for validation, and 11,405 for testing. It features annotations like polygonal segmentation and bounding boxes for figures, lists, titles, tables, and texts in images sourced from research papers and articles. The dataset's evaluation employed the COCO analytics method \cite{coco14}. We selectively used 102,514 images from the 86,460 table annotations in PubLayNet for our experiments.

\noindent\textbf{PubTables:} PubTables-1M~\cite{pubtables5}, specifically tailored for table detection in scientific documents, is an extensive dataset featuring nearly one million tables. It stands out for its comprehensive annotations, including precise location information, crucial for accurately detecting tables within diverse documents. Its large scale and meticulous annotations make it a significant resource for developing and refining table detection algorithms.

\noindent\textbf{ICDAR-19:} The ICDAR 2019 competition for Table Detection and Recognition (cTDaR) \cite{icdar19} introduced two novel datasets (modern and historical) for the table detection task (TRACK A). To facilitate direct comparisons with previous methods \cite{Ayan29}, we provide results at an Intersection over Union (IoU) threshold of 0.8 and 0.9.

\subsection{Evaluation Criteria}
\label{sec:Eval}
We assess the effectiveness of our semi-supervised table detection method through specific evaluation metrics: Precision, Recall, and F1-score. Precision \cite{pr61} is the ratio of correctly predicted positive observations (True Positives) to the total predicted positive observations (True Positives + False Positives). Recall \cite{pr61} measures the proportion of actual positives correctly identified (True Positives) out of the total actual positives (True Positives + False Negatives). The F1-score \cite{pr61} is the harmonic mean of Precision and Recall. Moreover, We evaluate our approach using AP@50 and AP@75, which assess precision at 50\% and 75\% IoU thresholds, reflecting moderate and high localization accuracy respectively, alongside average recall, measuring our model's capacity to detect all relevant instances

\subsection{Implementation Details}
\label{sec:implement}
We use the ResNet-50 backbone on 8 Nvidia RTXA6000 GPUs, initially trained on the ImageNet dataset, to evaluate the effectiveness of our semi-supervised method. We train on a diverse range of datasets, including PubLayNet, ICDAR-19, PubTables, and all subsets of the TableBank dataset, taking randomly 10\%, 30\%, and 50\% labeled data with the remaining as unlabeled. We conduct pseudo-labeling with a 0.7 threshold and optimize using AdamW. Our training spans 120 epochs, reducing the learning rate by 10\% after the 110th epoch, and we typically set our batch size to 16.
We adopt DETR's data augmentation strategy, which involves horizontal flipping, random cropping, and resizing. Additionally, we apply strong augmentation techniques such as horizontal flips, resizing, patch removal, cropping, conversion to grayscale, and Gaussian blur. For weak augmentation, we focus mainly on horizontal flipping. Setting the number of queries (N) in the decoder to 30 gives the best results. Our resizing approach ensures the image's longest side is at most 1333 pixels and the shortest side is at least 480 pixels. These strategic adjustments and augmentations boost the model's performance and efficiency.
\begin{table}
\begin{minipage}[b]{.51\textwidth}
\begin{center}
\caption{Performance of our semi-supervised transformer-based approach on different splits of TableBank dataset with varying percentage label data. }\label{tab:tablebank}
\renewcommand{\arraystretch}{1} 
\begin{tabular*}{\textwidth}
{@{\extracolsep{\fill}}llllll@{\extracolsep{\fill}}}
\toprule
\textbf{Dataset} &
\textbf{Labels} &
\textbf{mAP} & 
\textbf{AP\textsuperscript{50}} &
\textbf{AP\textsuperscript{75}}  & 
\textbf{AR\textsubscript{L}}  \\
\toprule
\multirow{3}{*}{TableBank-word }  & 10$\%$  & 92.9 & 95.3 & 93.9 & 97.4  \\
& 30$\%$ & 94.1 & 95.8 & 94.5 & 98.2  \\
& 50$\%$ & 94.3 & 95.8 & 94.8 & 98.3 \\
\midrule
 \multirow{3}{*}{TableBank-latex } & 10$\%$ & 91.2 & 97.6 & 96.4 & 95.3 \\
 & 30$\%$ & 93.7  & 97.3 & 96.3 & 97.7  \\
 & 50$\%$ & 94.8 & 97.9 & 97.0 & 98.1 \\
 \midrule
\multirow{3}{*}{TableBank-both }  & 10$\%$  & 92.7 & 95.8 & 94.6 &  93.6   \\
& 30$\%$ & 93.8 & 95.2 & 95.2 & 93.6 \\
& 50$\%$ & 94.2 & 96.1 & 95.8 & 95.8 \\
\bottomrule
\end{tabular*}
\end{center}
\end{minipage}
\hspace{10pt}
\begin{minipage}[b]{.45\textwidth}
\begin{center}
\caption{Recall results comparison of our semi-supervised approach with previous semi-supervised table detection approach. Here Def-semi refers to \cite{shehzadi_semi-detr_table}. }\label{tab:tablebank_AR}
\renewcommand{\arraystretch}{1} 
\begin{tabular*}{\textwidth}
{@{\extracolsep{\fill}}llll@{\extracolsep{\fill}}}
\toprule
\textbf{Dataset} &
\textbf{Labels} &
\textbf{Def-semi} & 
\textbf{Our}   \\
\toprule
\multirow{3}{*}{TableBank-word }  & 10$\%$  & 87.1 & 97.4  \\
& 30$\%$ & 92.1 & 98.2  \\
& 50$\%$ & 94.5 & 98.3 \\
\midrule
 \multirow{3}{*}{TableBank-latex } & 10$\%$ & 74.3 & 95.3 \\
 & 30$\%$ & 89.0 & 97.7  \\
 & 50$\%$  & 91.4 & 98.1 \\
 \midrule
\multirow{3}{*}{TableBank-both }  & 10$\%$ & 90.1 &  93.6   \\
& 30$\%$ & 91.5 & 93.6 \\
& 50$\%$ & 95.3 & 95.8 \\
\bottomrule
\end{tabular*}
\end{center}
\end{minipage}
\end{table}
\section{Results and Discussion}
\label{sec:results}
\subsection{TableBank}
In our study, we evaluate our approach using the TableBank dataset, examining performance across various splits with different proportions of labeled data: 10\%, 30\%, and 50\%. Table~\ref{tab:tablebank} shows we achieve mAP of 92.9\%, 91.2\%, and 92.7\% by using 10\% labels of TableBank word, latex, and both splits, respectively.
Unlike previous semi-supervised table detection method~\cite{shehzadi_semi-detr_table}, which employs deformable DETR~\cite{Deformable54} with a focus on improving the attention mechanism to improve the performance. Our semi-supervised approach optimizes the matching process between object queries and image features. As a result, our semi-supervised strategy achieves significantly higher recall rates than earlier semi-supervised methods, as shown in Tables~\ref{tab:tablebank_AR}. This improvement shows the effectiveness of semi-supervised table detection, particularly when dealing with limited labeled data.
\vspace{-1pt}
\begin{table}
\begin{center}
\caption{Comparative analysis of our semi-supervised approach with previous supervised and semi-supervised methods on the TableBank-Both dataset using 10\%, 30\%, and 50\% labeled data.  Here, the results are reported on mAP.}\label{tab:comptablebank}
\begin{tabular*}{.9\textwidth}
{@{\extracolsep{\fill}}cccccc@{\extracolsep{\fill}}}
\toprule
\textbf{Method} & 
\textbf{Approach} &
\textbf{Detector} &
\textbf{$10\%$ } &
\textbf{$30\%$ }  & 
\textbf{$50\%$ }  \\
\toprule
 Ren et al. \cite{faster23}  & supervised  & Faster R-CNN & 80.1 & 80.6 & 83.3\\

Zhu et al. \cite{Deformable54}  & supervised  & Deformable DETR & 80.8 & 82.6 & 86.9\\

STAC \cite{simplesemi76} &   semi-supervised & Faster R-CNN  & 82.4 & 83.8 & 87.1   \\

Unbiased Teacher \cite{unbiasedT36} &   semi-supervised & Faster R-CNN  & 83.9 & 86.4 & 88.5\\

Humble Teacher \cite{humbleTeacher6} &   semi-supervised & Faster R-CNN &  83.4 & 86.2 & 87.9 \\

Soft Teacher \cite{softTeacher56} &   semi-supervised & Faster R-CNN  & 83.6 & 86.8 &  89.6 \\

Shehzadi et al.~\cite{shehzadi_semi-detr_table} & semi-supervised  & Deformable DETR  & 84.2 & 86.8 & 91.8 \\

Our &   semi-supervised  & Sam-DETR  & \textbf{92.7} & 
\textbf{93.8} & \textbf{94.2} \\
\bottomrule
\end{tabular*}
\end{center}
\end{table}
Table~\ref{tab:comptablebank} presents a comparative analysis of our semi-supervised approach against prior supervised and semi-supervised methods using the TableBank-both dataset, which includes splits with 10\%, 30\%, and 50\% labeled data. The outcomes demonstrate that our approach outperforms the earlier methods across these varying levels of labeled data. This is a significant finding, highlighting the effectiveness of our semi-supervised strategy in scenarios with limited labeled data availability.

\begin{figure}
\centering
\includegraphics[width=0.95\textwidth]{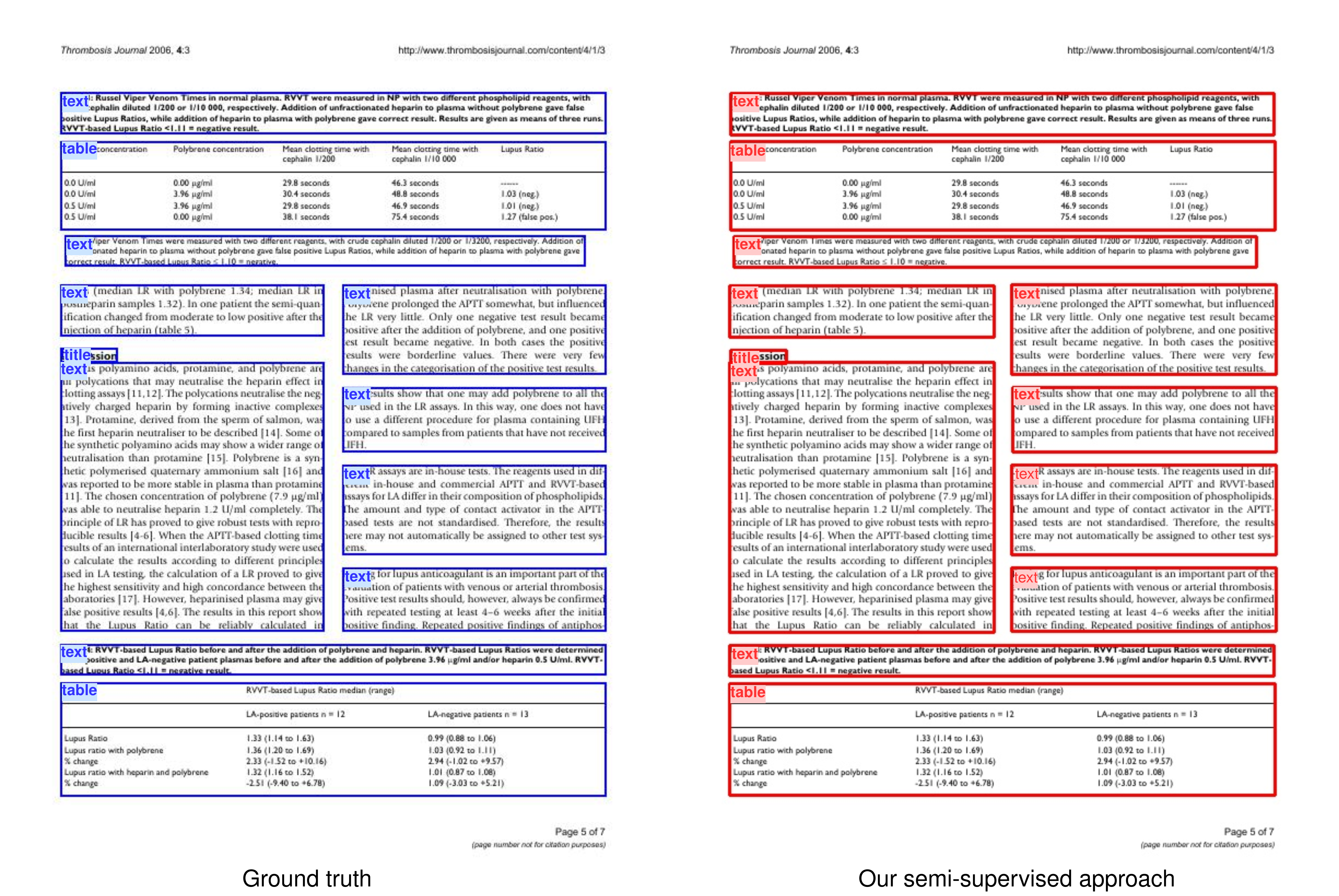}
\caption{Visual Analysis of our semi-supervised approach. Here, blue represents ground truth and red denotes our predictions results using 10\% labels on PubLayNet datatset.}\label{fig:semi-results}
\end{figure}
\subsection{PubLayNet}
We also evaluate the performance of our transformer-based semi-supervised learning model on the PubLayNet dataset, experimenting with different ratios of labeled to unlabeled data (10\%, 30\%, and 50\%). This study aims at understanding the model's performance in scenarios with limited labeled data, a common challenge in real-world applications.
Table~\ref{tab:publaynet} shows we achieve mAP of 89.9\%, 90.9\%, and 93.2\% by using 10\%, 30\%, and 50\% labels of PubLayNet dataset. We shows the visual analysis of our semi-supervised approach in Fig.~\ref{fig:semi-results}. Our semi-supervised approach also provides higher recall than the previous semi-supervised approach, as observed in Table~\ref{tab:publaynet_AR}.
\vspace{-10pt}
\begin{table}
\begin{minipage}[b]{.54\textwidth}
\caption{Performance of our semi-supervised transformer-based approach on PubLayNet dataset with varying percentage label data.}\label{tab:publaynet}
\renewcommand{\arraystretch}{1} 
\begin{tabular*}{\textwidth}
{@{\extracolsep{\fill}}ccllll@{\extracolsep{\fill}}}
\toprule
\textbf{Dataset} & 
\textbf{Label-percent} &
\textbf{mAP} & 
\textbf{AP\textsuperscript{50}} &
\textbf{AP\textsuperscript{75}}  & 
\textbf{AR\textsubscript{L}}  \\
\toprule
\multirow{3}{*}{PubLayNet} & 10\% & 89.9 & 97.1 & 94.3 & 96.6  \\

& 30\% & 90.9 & 97.4 & 94.9 & 96.9  \\

 & 50\% & 93.2 & 97.7 & 95.0 & 97.3  \\
\bottomrule
\end{tabular*}
\end{minipage}\qquad
\begin{minipage}[b]{.4\textwidth}
\begin{center}
\caption{Recall results comparison of our approach with previous semi-supervised table detection approach.} \label{tab:publaynet_AR}
\renewcommand{\arraystretch}{1} 
\begin{tabular*}{\textwidth}
{@{\extracolsep{\fill}}cccc@{\extracolsep{\fill}}}
\toprule
\textbf{Method} & 
\textbf{10\%} &
\textbf{30\%} &
\textbf{50\%} \\
\toprule
Shehzadi et al.~\cite{shehzadi_semi-detr_table}   & 91.0 & 93.2 & 96.0 \\

Our & \textbf{96.6} & \textbf{96.9} & \textbf{97.3} \\
\bottomrule
\end{tabular*}
\end{center}
\end{minipage}
\end{table}

We also compare our approach against traditional deep learning methods, both supervised and semi-supervised, to highlight advancements. A key focus is the model's performance with only 10\% labeled data, where we observe that our approach achieves the highest mAP score of 89.9, as detailed in Table~\ref{tab:comppublaynet}. This shows the effectiveness of our method in leveraging minimal labeled data, demonstrating the significant potential of our approach for practical applications in table detection and recognition.
\vspace{-1pt}
\begin{table*}[htp!]
\begin{center}
\caption{Comparative analysis of our semi-supervised approach with previous supervised and semi-supervised methods on PubLayNet table class dataset using 10\%, 30\%, and 50\% labeled data. Here, the results are reported on mAP.} \label{tab:comppublaynet}
\renewcommand{\arraystretch}{1} 
\begin{tabular*}{.8\textwidth}
{@{\extracolsep{\fill}}cccccc@{\extracolsep{\fill}}}
\toprule
\textbf{Method} & 
\textbf{Approach} &
\textbf{Detector} & 
\textbf{$10\%$ } &
\textbf{$30\%$ }  & 
\textbf{$50\%$ }  \\
\toprule
 Ren et al. \cite{faster23} & supervised  & Faster R-CNN  & 83.4 & 86.6 & 87.9    \\

Zhu et al. \cite{Deformable54} & supervised  & Deformable DETR  & 83.9 & 86.8 & 88.1    \\

Soft Teacher \cite{softTeacher56} & semi-supervised & Faster R-CNN  & 88.3  & 89.5 & 92.5  \\

Shehzadi et al.~\cite{shehzadi_semi-detr_table} & semi-supervised  & Deformable DETR & 88.4 & 90.3 & 92.8 \\

Our & semi-supervised  & SAM-DETR & \textbf{89.9} & \textbf{90.9} & \textbf{93.2} \\
\bottomrule
\end{tabular*}
\end{center}
\end{table*} 
\vspace{-10pt}
\subsection{PubTables}
In this subsection, we detail our experimental results for the PubTables dataset in a semi-supervised setting using different percentages of labeled data. Our analysis includes a comparison between our transformer-based semi-supervised method and earlier CNN-based and transformer-based supervised approaches. As shown in Table~\ref{tab:PubTabless}, our semi-supervised approach achieves a 92.3 mAP score even with only 10\% of the data labeled, which highlights the effectiveness of our method in utilizing a smaller amount of labeled data to attain high accuracy.
\begin{table*}[htp!]
\begin{center}
\begin{minipage}[b]{.65\textwidth}
\caption{Performance of our semi-supervised transformer-based approach on the PubTables dataset with varying levels of labeled data (10\%, 30\%, 50\%). Results show high accuracy with even a minimal amount of labeled data.}\label{tab:PubTabless}
\renewcommand{\arraystretch}{1} 
\begin{tabular*}{\textwidth}
{@{\extracolsep{\fill}}ccllll@{\extracolsep{\fill}}}
\toprule
\textbf{Dataset} & 
\textbf{Label} &
\textbf{mAP} & 
\textbf{AP\textsuperscript{50}} &
\textbf{AP\textsuperscript{75}}  & 
\textbf{AR\textsubscript{L}}  \\
\toprule
\multirow{3}{*}{PubTables} & 10\% & 92.3 & 93.7 & 93.8 & 87.8   \\

                           & 30\% & 93.5 & 94.8 & 93.7 & 88.1   \\

                           & 50\% & 93.8 & 94.8 & 94.8 & 88.3 \\
\bottomrule
\end{tabular*}
\end{minipage}
\end{center}
\end{table*} 

Table~\ref{tab:pubtables-comp} presents a comparison between our semi-supervised approach and previous supervised methods. While a direct comparison isn't feasible due to different percentages of label data for training, our results are notably comparable. For instance, a Faster R-CNN model trained on fully labeled data achieved an mAP of 82.5, whereas our semi-supervised approach reached an mAP of 92.3 using only 10\% labeled data.
\begin{table*}[htp!]
\begin{center}
\begin{minipage}[b]{.75\textwidth}
\caption{Comparative Analysis of Semi-Supervised and Supervised Methods. It clearly shows that our semi-supervised model achieves comparable results even with limited data. } \label{tab:pubtables-comp}
\renewcommand{\arraystretch}{1.1} 
\begin{tabular*}{\textwidth}
{@{\extracolsep{\fill}}cccccc@{\extracolsep{\fill}}}
\toprule
\textbf{Method} & 
\textbf{Approach} &
\textbf{Detector} & 
\textbf{mAP} & 
\textbf{AP\textsuperscript{50}} &
\textbf{AP\textsuperscript{75}}   \\
\toprule
Smock et al.~\cite{pubtables5} & supervised & Faster R-CNN   & 82.5 & 98.5 & 92.7  \\
Smock et al.~\cite{pubtables5}  & supervised & DETR & 96.6 & 995 & 98.8 \\
Our & semi-supervised (10\%) & SAM-DETR & 92.3 & 93.7 & 93.8    \\
\bottomrule
\end{tabular*}
\end{minipage}
\end{center}
\end{table*} 

\noindent\textbf{Comparisons with Previous Table Detection Approaches.}
In Table~\ref{tab:compall}, we present a comprehensive comparison of our semi-supervised table detection approach against existing supervised and semi-supervised methods. 
\begin{table*}[htp!]
\begin{center}
\caption{Comparative analysis of our semi-supervised approach with previous supervised and semi-supervised methods. Here, the results are reported on mAP.} \label{tab:compall}
\renewcommand{\arraystretch}{1} 
\begin{tabular*}{\textwidth}
{@{\extracolsep{\fill}}cccccc@{\extracolsep{\fill}}}
\toprule
\textbf{Method} & 
\textbf{Approach} & 
\textbf{ Labels} &
\textbf{TableBank} &
\textbf{PubLayNet }  & 
\textbf{PubTables} \\
\toprule
 CDeC-Net \cite{Agarwal52} & supervised & 100\%  &  96.5 & 97.8 & -    \\
CasTabDetectoRS \cite{CasTab45} & supervised & 100\% & 95.3 & - & -     \\
Faster R-CNN \cite{PubLayNet3} & supervised & 100\% & - & 90 &     \\
VSR \cite{vsr45} & supervised & 100\% & - & 95.69 &    \\
Smock et al.~\cite{pubtables5}    & supervised & 100\%  & - & - &  96.6  \\
Shehzadi et al.~\cite{shehzadi_semi-detr_table} & semi-supervised & 10\%  & 84.2 & 88.4 & -   \\
Our & semi-supervised & 10\%  & 92.7 & 89.9 & 92.3   \\
\bottomrule
\end{tabular*}
\end{center}
\end{table*} 
Our approach facilitates learning with significantly fewer labeled instances. Our semi-supervised method performs well despite limited labeled data, achieving high mAP scores on datasets and outperforming previous semi-supervised models. It shows improved performance in scenarios with scarce labeled data, offering comparable results to fully supervised methods while using only 10\% of their labeled data.
\subsection{ICDAR-19}
In our analysis, we additionally conduct an evaluation of the ICDAR-19 TrackA table detection dataset across different Intersection over Union (IoU) thresholds using 50\% labeled data. Furthermore, we compare our semi-supervised approach with earlier supervised and semi-supervised strategies, as depicted in Table~\ref{tab:icdar19}. The results, utilizing 50\% labeled data, show that our transformer-based semi-supervised framework surpasses prior semi-supervised methods, demonstrating superior accuracy.
\begin{table*}[htp!]
\begin{center}
\caption{Performance comparison between the proposed 
semi-supervised approach and previous state-of-the-art results on the dataset of ICDAR 19 Track A (Modern). } \label{tab:icdar19}
\renewcommand{\arraystretch}{1} 
\begin{tabular*}{\textwidth}
{@{\extracolsep{\fill}}cccccccc@{\extracolsep{\fill}}}
\toprule
\textbf{Method} &
\textbf{Approach} &
&
\textbf{IoU=0.8} &
&& 
\textbf{IoU=0.9} & \\
\cline{3-8} 
& &\textbf{Recall} &
\textbf{Precision} &
\textbf{F1-Score} &
\textbf{Recall} &
\textbf{Precision} &
\textbf{F1-Score} \\
\toprule
 TableRadar~\cite{icdar19}  & supervised & 94.0 & 95.0 & 94.5 & 89.0 & 90.0 & 89.5    \\
NLPR-PAL~\cite{icdar19}  & supervised & 93.0 & 93.0 & 93.0 & 86.0 & 86.0 & 86.0    \\
Lenovo Ocean~\cite{icdar19}  & supervised & 86.0  & 88.0 & 87.0 & 81.0  & 82.0 & 81.5   \\
CDeC-Net~\cite{Agarwal52} & supervised & 93.4  & 95.3 & 94.4 & 90.4 & 92.2 & 91.3    \\
HybridTabNet~\cite{Hyb65} & supervised & 93.3 & 92.0 & 92.8 & 90.5 & 89.5 & 90.2    \\
Shehzadi et al.~\cite{shehzadi_semi-detr_table} & semi-supervised (50\%)  & 71.1 & 82.3 & 76.3 & 66.3 & 76.8 & 71.2  \\
Our & semi-supervised (50\%)  & 73.5 & 83.8 & 77.2 & 68.4 & 77.8 & 72.1  \\
\bottomrule
\end{tabular*}
\end{center}
\end{table*} 
\section{Ablation Study}
In the ablation study, we evaluate the model's performance using only 30\% of the labeled data from the PubTables dataset. The study observes the effect of varying the pseudo-labeling confidence threshold, the number of filtered pseudo-labels, and the number of learnable queries, offering insights into their roles in enhancing model performance in document analysis tasks.

\noindent\textbf{Pseudo-Labeling confidence threshold}
The choice of a confidence threshold in pseudo-labeling influences the performance of our semi-supervised approach, as observed in Table~\ref{tab:filter}.
\begin{table}
\begin{minipage}[b]{.49\textwidth}
\caption{Performance comparison using different Pseudo-labeling confidence threshold values. The best threshold values are shown in bold.}\label{tab:filter}
\renewcommand{\arraystretch}{0.7} 
\begin{tabular*}{1\textwidth}
{@{\extracolsep{\fill}}cccc@{\extracolsep{\fill}}}
\toprule
\textbf{Threshold} & 
\textbf{AP} & 
\textbf{AP\textsuperscript{50}} &
\textbf{AP\textsuperscript{75}}    \\
\toprule
0.5 & 89.8 & 91.3 & 90.4  \\
0.6 &  90.4 & 92.1 & 91.5  \\
\textbf{0.7}  & \textbf{93.5} & \textbf{94.8} &  \textbf{93.7} \\

0.8  & 90.2 & 91.7 & 90.2  \\

0.9 & 88.6 & 89.3 & 89.1  \\
\bottomrule
\end{tabular*}
\end{minipage}\qquad
\begin{minipage}[b]{.48\textwidth}
\begin{center}
\caption{Performance comparison using different numbers of learnable queries to the decoder input. Here, the best performance results are shown in bold.
}\label{tab:queries}
\renewcommand{\arraystretch}{0.7} 
\begin{tabular*}{\textwidth}
{@{\extracolsep{\fill}}cccc@{\extracolsep{\fill}}}
\toprule
\textbf{Queries} & 
\textbf{AP} & 
\textbf{AP\textsuperscript{50}} &
\textbf{AP\textsuperscript{75}}     \\
\toprule
 10 &  88.5 & 87.8 & 86.8   \\
\textbf{30} & \textbf{93.5} & \textbf{94.8} &  \textbf{93.7} \\
60 &  91.8 & 92.8 & 91.5    \\
100 & 88.6 & 90.2 & 87.3   \\
300 & 82.1 & 85.3 & 84.1  \\
\bottomrule
\end{tabular*}
\end{center}
\end{minipage}
\end{table}
A low threshold leads to the filtering of a large number of pseudo-labels. However, these include incorrect pseudo-labels, introducing noise into the training process, and potentially degrading the model's performance. On the other hand, a high threshold ensures the generation of high-quality pseudo-labels, reducing the risk of noise. However, this results in fewer pseudo-labels fed into the student network, thus not fully leveraging the advantages of semi-supervised learning. The balance between generating enough pseudo-labels and ensuring that these pseud-labels are accurate enough to be useful is crucial in optimizing model performance.
 
\noindent\textbf{Influence of Learnable queries Quantity}
We examine the effect of both increasing and decreasing the number of input queries on the performance of our semi-supervised approach, as highlighted in Table~\ref{tab:queries}. While increasing the queries can improve the model's ability to detect and focus on a wide range of features, enhancing accuracy in complex detection tasks, it also leads to more overlapping predictions, necessitating the use of Non-Maximum Suppression (NMS). Conversely, decreasing the number of queries reduces computational complexity but limits the model's detection capabilities. We find that our model achieves the best performance with 30 queries. Deviating from this optimal count, whether by increasing or decreasing the number of queries, significantly impacts the model's accuracy and efficiency.
\vspace{-15pt}
\begin{table}[ht]
\begin{center}
\caption{Performance evaluation using top-k pseudo-labels. The best results are in bold.}
\label{tab:topk}
\renewcommand{\arraystretch}{1} 
\begin{tabular*}{.5\textwidth}
{@{\extracolsep{\fill}}cccc@{\extracolsep{\fill}}}
\toprule
\textbf{Top-k} & 
\textbf{AP} & 
\textbf{AP\textsuperscript{50}} &
\textbf{AP\textsuperscript{75}}  \\
\toprule
 1 & 90.5 & 93.8 & 91.2    \\
 2 & 91.7 & 94.4 & 91.9    \\
\textbf{3} & \textbf{93.5} & \textbf{94.8} & \textbf{93.7}   \\
 4 & 92.8  & 94.2 & 92.5 \\
\bottomrule
\end{tabular*}
\end{center}
\end{table}

\noindent\textbf{Influence of quantity of Pseudo-label Filtering}
In Table~\ref{tab:topk}, we observe the impact of varying quantities of filtered pseudo-labels generated by the teacher network on model performance. While including more pseudo-labels enhances model performance, it is also vital to consider their quality. Selecting more pseudo-labels, such as the top-4, inherently introduces some lower-quality labels into the training process. Including less reliable pseudo-labels can adversely affect the model's performance, highlighting the need for a balanced approach in pseudo-label selection that optimizes quantity and quality to achieve the best model performance.
\section{Conclusion}
\label{sec:conclusion}
Our research addresses the challenge of accurately and efficiently detecting document objects, such as tables and text, in semi-supervised settings. This approach utilizes minimal labeled data and employs student-teacher networks that mutually update during training. Previous transformer-based research focused on improving attention or increasing the number of object queries, which impacts training time and performance. We eliminate the need for NMS and focus on matching between object queries and image features. Our novel approach using SAM-DETR in a semi-supervised setting helps align object queries with target features, significantly reducing false positives and improving the detection of document objects in complex layouts. In short, our semi-supervised method enhances the accuracy of document analysis, particularly in scenarios with limited labeled data.

%
%
%
\bibliographystyle{IEEEtran}
\bibliography{main}

\end{document}